# Learning to Order Facts for Discourse Planning in Natural Language Generation

**Aggeliki Dimitromanolaki**
Department of Information &
Communication Systems Engineering
University of the Aegean

Institute of Informatics &
Telecommunications
NCSR "Demokritos"
15310, Ag.Paraskeui, Greece
`adimit@iit.demokritos.gr`

**Ion Androutsopoulos**

Department of Informatics
Athens University of Economics &
Business
Patission 76, 10434, Athens, Greece
`ion@aueb.gr`

## Abstract

This paper presents a machine learning approach to discourse planning in natural language generation. More specifically, we address the problem of learning the most natural ordering of facts in discourse plans for a specific domain. We discuss our methodology and how it was instantiated using two different machine learning algorithms. A quantitative evaluation performed in the domain of museum exhibit descriptions indicates that our approach performs significantly better than manually constructed ordering rules. Being retrainable, the resulting planners can be ported easily to other similar domains, without requiring language technology expertise.

## 1 Introduction

Along the lines of Reiter and Dale (2000), we view natural language generation (NLG) as consisting of six tasks: content determination, discourse planning, aggregation, lexicalization, referring expression generation, and linguistic realization. This paper is concerned with the second task, i.e., *discourse planning*. Discourse planning determines the ordering and rhetorical relations of the logical messages, hereafter called *facts*, that the generated document is intended to convey. Most existing approaches to discourse planning are based on either rhetorical structure theory (RST) (Mann and Thompson, 1988; Hovy, 1993) or schemata (McKeown, 1985). In both cases, the rules that determine the order and the rhetorical relations are typically written by hand. This is a time-consuming process, which requires domain and linguistic expertise, and has to be repeated whenever the system is ported to a new domain; see also Rambow (1990).

This paper presents a machine learning (ML) approach to the subtask of discourse planning that attempts to find the most natural ordering of facts in each generated document. Our approach was motivated by experience obtained from the M-PIRO project (Androutsopoulos et al., 2001). Building upon ILEX (O'Donnell et al., 2001), M-PIRO is developing technology that allows personalized descriptions of museum exhibits to be generated in several languages, starting from symbolic, language-independent information stored in a database, and small fragments of text (Isard et al., 2003). One of M-PIRO's most ambitious goals is to develop authoring tools that will allow domain experts, e.g., museum curators, with no language technology expertise to configure the system for new application domains. While this goal has largely been achieved for resources such as the domain-dependent parts of the ontology, or domain-dependent settings that affect content selection, lexicalization, and referring expression generation (Androutsopoulos et al., 2002), designing tools that will allow domain experts to edit discourse planning rules has proven difficult. In contrast, domain experts, in our case museum curators, were happy to reorder the clauses of sample generated texts, thus indicating the preferred orderings of the facts in the corresponding discourse plans. We have, therefore, opted for a machine learning approach that allows fact-ordering rules to be captured automatically from sets of manually reordered facts. We view this approach as a first step towards learning richer discourse plans, which apart from ordering information will also include rhetorical relations, although the experience from M-PIRO indicates that even just ordering the facts in a natural way can lead to quite acceptable texts. Being automatically retrainable, the planners that our approach produces can be easily ported to other similar domains, e.g., descriptions of products for e-commerce catalogues, provided that samples of ideal fact orderings can be made available.

Our method introduces a new representation of the fact-ordering task, and employs supervised learning algorithms. It is assumed that the number of facts to be conveyed by each generated document, in effect the desired length of the generated texts, has been fixed to a particular value; i.e., all the documents contain the same number of facts. In ILEX and M-PIRO, this number is provided by the user model. Furthermore, it is assumed that a content determination module is available, which selects the particular facts to be conveyed by each document. Our method consists of a sequence of stages, the number of stages being equal to the number of facts to be conveyed by each document. Each stage is responsible for the selection of the fact to be placed at the corresponding position in the resulting document. In our experiments, we set the number of facts per document to six, which per document to six, which seems to be an appropriate value for our particular domain and an average adult user, but this number could vary depending on the application and user type. Two learning algorithms, decision trees (Quinlan, 1993) and instance-based learning (Aha and Kibler, 1991), were explored. The results are compared against two baselines: a simple hand-crafted planner, which always assigns a predefined order, and the majority scheme. The latter selects, among the facts that are available at each position, the fact that occurred most frequently at that position in the training data. Overall, the results indicate that with either of the two learning algorithms our method significantly outperforms both of the baselines, and that there is no significant difference in the performance of the two learning algorithms.

The remainder of this paper is organized as follows. Section 2 presents previous learning approaches to NLG, and discusses their relevance to the work presented here. Section 3 describes our learning approach, including issues such as data representation and system architecture. Section 4 discusses our experiments and their results. Section 5 concludes and highlights plans for future work.

## 2  Previous work

In recent years, ML approaches have been introduced to NLG to address problems such as the construction and maintenance of domain and language resources, which is a time-consuming process in systems that use hand-crafted rules.[1] To the best of our knowledge, only two of these approaches (Duboue and McKeown, 2001; Duboue and McKeown, 2002) consider discourse planning.

Duboue and McKeown (2001) present an unsupervised ML algorithm based on pattern matching and clustering, which is used to learn ordering constraints among facts. The same authors have also used evolutionary algorithms to learn the tree representation of a planner (Duboue and McKeown, 2002). These works are similar to ours in that we also address the problem of ordering facts. However, Duboue and McKeown follow the lines of schema-based planning, where content determination is not an independent stage, but is interleaved with discourse planning. This means that the discourse planner has the overall control of content determination, and cannot handle inputs from an independent content determination module. In contrast, our method can be used with any content determination mechanism that returns a fixed number of facts. This has the benefit that alternative content determination modules can be used without affecting the discourse planner. Moreover, while Duboue and McKeown (2002) learn a tree structure representing the best sequence of facts, our method directly manipulates facts.

---

[1] For an extensive bibliography on statistical and machine learning approaches to NLG, see:
http://www.iit.demokritos.gr/~adimit/bibliography.html.

Mellish et al. (1998) also experiment with genetic algorithms to find the optimal RST tree, which is then mapped to the corresponding sequence of facts. Karamanis and Manurung (2002) use a similar approach that employs constraints from Centering Theory in the genetic search. However, these approaches do not involve any learning: the genetic search is repeated every time the text planner is invoked, i.e., for each new document. In contrast, our method induces a single discourse planner from the training data, which is then used to order any set of facts provided by the content determinator.

ML approaches to NLG have also been used in syntactic and lexical realization (Langkilde and Knight, 1998; Bangalore and Rambow, 2000; Ratnaparkhi, 2000; Varges and Mellish, 2001; Shaw and Hatzivassiloglou, 1999; Malouf 2000), as well as in sentence planning tasks (Walker et al., 2001; Poesio et al., 2000). In the context of spoken dialogue systems, learning techniques have been used to select among different templates (Oh and Rudnicky, 2000; Walker, 2000). These approaches, however, are not directly relevant to discourse planning.

The problem of ordering semantic units has also been addressed in the context of summarization. Kan and McKeown (2002) use an n-gram model to infer ordering constraints between facts, while Barzilay et al. (2002) manually identify constraints on ordering, using a corpus of ordering preferences among subjects and clustering techniques that identify commonalities among these preferences. The approach presented here, instead of identifying ordering constraints, "learns" the overall ordering of the input facts.

## 3 Learning to order facts

In our approach, the discourse planner is trained on manually ordered sequences of facts of a fixed length. Once trained, it is able to determine what it considers to be the most natural ordering of any set of facts, as output by a content determination module, provided that the cardinality of the set is the same as the length of the training sequences. This section describes our approach in more detail, starting from the required data and the pre-processing that they undergo.

### 3.1 Data and pre-processing

Our data was derived from the database of M-PIRO. This database currently contains information about 50 museum exhibits, each of which is associated with a large number of facts. For example, the left column of Table 1 shows the database facts associated with the entity *exhibit9*. Each generated document is intended to describe a museum exhibit. As already mentioned, in our experiments the number of facts to be conveyed by each document was set to six. That is, when asked to describe *exhibit9*, the content determination module would choose six of the facts in the left column of Table 1, possibly depending on user modeling information, such as the interests and backgrounds of the users, or information indicating which facts have already been conveyed to the users. We did not use a particular content determination module, because we wanted the discourse planner to be independent from the content determination process. Our goal was to be able to order any set of six facts that could be provided as input by an arbitrary content determination module.

| Database facts | Selected facts (input to discourse planner) |
|---|---|
| subclass(EXHIBIT9,RHYTON)<br>current-location(EXHIBIT9,MUS-DU-PETIT-PALAIS)<br>original-location(EXHIBIT9,ATTICA)<br>potter-is(EXHIBIT9,SOTADES)<br>exhibit-characteristics(EXHIBIT9,ENTITY-1796)<br>painted-by(EXHIBIT9,PAINTER-OF-SOTADES)<br>creation-time(EXHIBIT9,DATE-1767)<br>creation-period(EXHIBIT9,CLASSICAL-PERIOD)<br>painting-technique-used(EXHIBIT9,RED-FIG-TECHN)<br>exhibit-depicts(EXHIBIT9,ENTITY-1786)<br>opposite-technique(RED-FIG-TECHN,BLACK-FIG-TEC)<br>technique-description(RED-FIG-TECHN,ENTITY-2474)<br>person-information(SOTADES,ENTITY-2739)<br>museum-country(MUS-DU-PETIT-PALAIS,FRANCE)<br>period-story(CLASSICAL-PERIOD,STORY-NODE4019) | f1: subclass(EXHIBIT9,RHYTON)<br>f2: current-location(EXHIBIT9,MUS-DUPETITPALAIS)<br>f3: original-location(EXHIBIT9,ATTICA)<br>f4: painted-by(EXHIBIT9,PAINTER-OF-SOTADES)<br>f5: creation-time(EXHIBIT9,DATE-1767)<br>f6: creation-period(EXHIBIT9,CLASSICAL-PERIOD) |

**Table 1: Database facts and facts selected as input to the discourse planner**

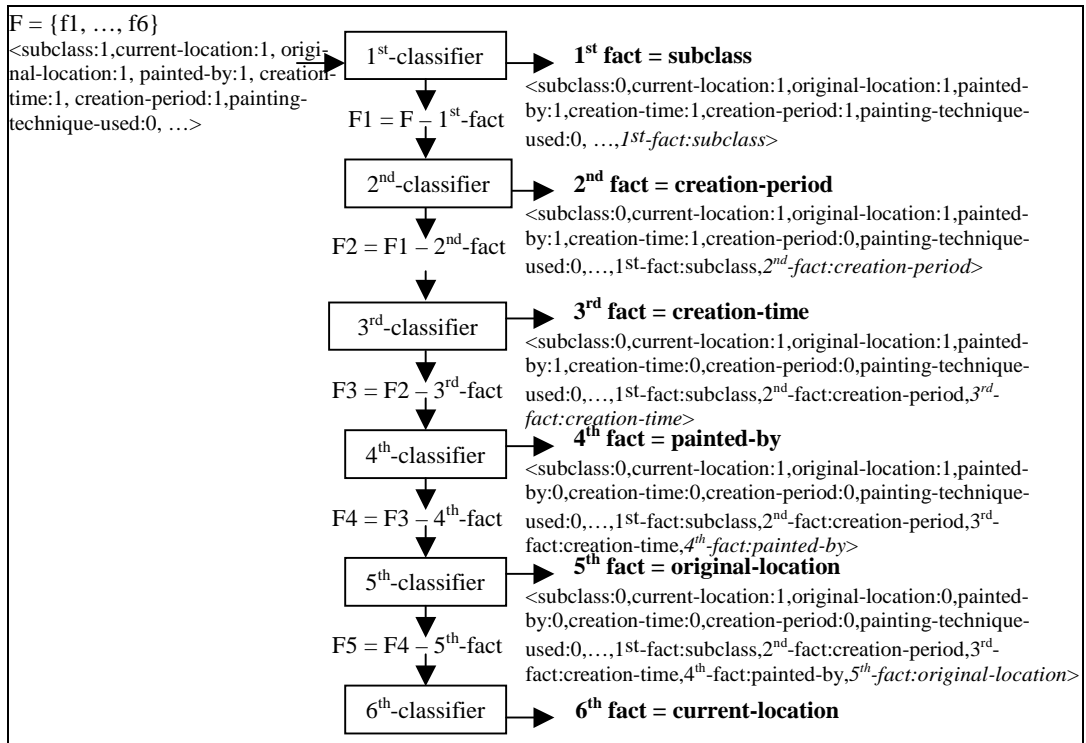

Figure 1: Architecture diagram

In order to create the dataset of our experiments, we used a program that yields all the possible combinations of six facts for each exhibit. The right column of Table 1 shows an example set of six facts, which can be used as input to the discourse planner. Many combinations, however, looked unreasonable in our domain; e.g., combinations that do not include the *subclass* fact (descriptions in the museum domain must always inform the reader about the type of the exhibit), or combinations that include facts providing background information about an entity that is not present in the discourse (for instance, combinations that include *opposite-technique* but not *painting-*

*technique-used* in Table 1). A refinement operation was performed manually to discard such combinations. We note that in real-life applications, the combinations would be obtained by calling several times a content determination module; hence, no refinement operation would be necessary, as the content determination module would, presumably, never return unreasonable combinations of facts.

After the refinement operation, 880 combinations of 6 facts were left. The facts of each set were manually assigned an order, to reflect what a domain expert considered to be the most natural ordering of the corresponding clauses in the generated texts. Each one of the 880 sets was then used as an instance in the learning algorithms, as will be explained in the following section.

## 3.2 Instance representation and planner architecture

Figure 1 shows the discourse planning architecture that our approach adopts, along with an example of inputs and outputs at each stage. We decompose the fact-ordering task into six multi-class classification problems. Each of the six classifiers selects the fact to be placed at the corresponding position. Each input set of six facts is represented as a vector in a multi-dimensional space, where dimensions correspond to values of attributes. 42 binary attributes, representing the fact types of the domain, were used. The vector at the top left corner of Figure 1 represents the set of six facts of the right column of Table 1. Each attribute shows whether a particular fact type exists in the input (e.g., creation-period:1) or not (e.g., painting-technique-used:0). Classifiers 2–6 have additional attributes representing the fact types that have already been selected for positions 1–5. More specifically, as shown in Figure 1, the attribute $1^{st}$-fact is added from the $2^{nd}$ classifier onwards, the attribute $2^{nd}$-fact is added from the $3^{rd}$ classifier onwards, and so forth. Therefore, the classifiers make their decisions based on the fact types that are present in their inputs (set of remaining facts to be ordered) and the fact types that have been selected at the previous positions. We assume that it is not possible to have more than one fact of the same type in the input set of facts because this is the case in the M-PIRO domain (e.g., we cannot have two facts of type *creation-period*) as well as in other similar domains. In a more general case, however, one could differentiate between facts of the same type, by enriching, for instance, the attributes, so as to represent information about the entities related with each fact, or by adding new attributes.

The output of each classifier is the class value representing the fact type that has been selected for the corresponding position. In the example of Figure 1, the classifiers select the following order: *subclass, creation-period,*

*creation-time, painted-by, original-location, current-location.* As shown in Figure 1, the sixth classifier has no substantial role, since there is only one fact left in the input, and, consequently, this fact will be placed at the sixth position.

In a similar manner, a sequence of $n$ classifiers can be used when each document is to convey $n$, rather than 6, facts. A limitation of this approach is that it cannot be used when $n$ varies across the documents. However, this is not a problem in M-PIRO, where $n$, in effect the length of the documents, is fixed for each user type: if there are $t$ user types, we train $t$ different document planners, one for each user type; each planner is a sequence of $n_i$ classifiers, where $n_i$ is the value of $n$ for the corresponding user type ($i = 1, …, t$).

## 4 Experiments and results

In order to evaluate our approach, we performed four experiments. The first experiment was conducted using the majority scheme, where each classifier selects among the available classes (i.e., among the facts that are present in the input set and have not been selected by the previous classifiers) the class (i.e., fact) that was most frequent in its training data. However, this scheme is too primitive, and could not be seen as a safe benchmark for our experiments. For this reason, we constructed a simple planner, hereafter *base planner*, which always assigns a predefined fixed order defined in collaboration with a museum expert; e.g., *subclass* should always be placed before *creation-period*, *creation-period* should always be placed before *creation-time*, etc. The base planner was used as a second baseline. In this way, we had a safer benchmark for the performance of the learning schemes. In the two remaining experiments we used instance-based and decision-tree learning. More specifically, we experimented with the $k$-nearest neighbour algorithm (Aha and Kibler, 1991), with $k = 1$, and the C4.5 algorithm (Quinlan, 1993). All the experiments were performed using the machine learning software of WEKA (Witten and Frank, 1999).

Figure 2 presents the accuracy scores of each of the six classifiers, for each learning scheme. The results were obtained using *10-fold cross-validation*. That is, the dataset (880 vectors) was divided into ten disjoint parts (folds), and each experiment was repeated 10 times. Each time, a different part was used for testing, and the remaining 9 parts were used for training. The dataset was *stratified*, i.e. the class distribution in each fold was approximately the same as in the full dataset. The reported scores are averaged over the 10 iterations. Accuracy measures the percentage of correct selections at each classifier (position) compared to the selections made by the human annotator. All schemes have 100% accuracy at the selection of the $1^{st}$ and $6^{th}$ fact. This happens because the first classifier always selects the fact *subclass*, which is always the first fact in our domain, while the sixth classifier has no alternative choice, since only one fact has been left in the input. At the other positions, both C4.5 and 1-NN perform better than the two baselines; C4.5 seems to have a slightly better performance than 1-NN. Paired two-tailed *t*-tests at p = 0.005 indicate that the observed differences in accuracy between baselines and ML schemes are statistically significant; the only exception is the selection of the $2^{nd}$ fact, where there is no significant difference between the base planner and 1-NN.

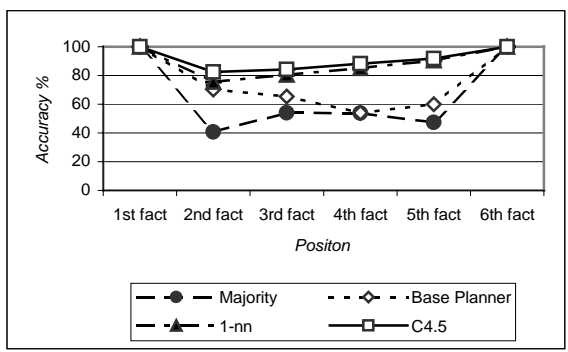

**Figure 2: Accuracy scores at each classification stage**

Figure 3 shows a text corresponding to the ordering produced by C4.5. The surface text, including aggregation and referring expression generation, was generated by hand, though we plan to automate this process using the corresponding modules of M-PIRO. The ordering of the facts, which are realized as natural language clauses, looks quite reasonable. The flow of information is not the optimal one, but does not cause problems to the understandability or readability of the text. Figure 4 shows the text that corresponds to the ordering of the human annotator. The two texts differ in the placement of the fact *made-of*, which is expressed as "it is made of marble"; C4.5 places this fact at the fourth position instead of the second, which is the right position according to the human annotator. The word "but" in the human text of Figure 4 implies the use of a rhetorical relation; the presence of this relation suggests a possible explanation of why the human text is ordered differently than the one produced by the system. The misplaced fact is penalized three times when computing the accuracy scores of the six classifiers: at the second classifier, where the fact *exhibit-portrays* is selected instead of *made-of*, at the third classifier, where *creation-period* is selected instead of *exhibit-portrays*, and at the fourth classifier, where *made-of* is selected instead of *creation-period*. This implies that the accuracy scores that were presented above are a very strict measure of the performance of our method, and, in fact, our method may actually be performing even better than what the scores indicate.

This exhibit is a portrait. It portrays Alexander the Great and was created during the Hellenistic period. It is made of marble. What we see in the picture is a roman copy. Today it is located at the archaeological museum of Thassos.

**Figure 3: Ordering of facts produced using C4.5**

This exhibit is a portrait. It is made of marble and portrays Alexander the Great. It was created during the Hellenistic period, but what we see in the picture is a roman copy. Today it is located in the archaeological museum of Thassos.

**Figure 4: Ordering of facts as specified by the human annotator**

We are currently trying to devise evaluation measures that are better suited to discourse planning, and to NLG in general. More specifically, we plan to apply metrics that assign different penalties depending on the importance of an error, based on the edit distance between the output of the discourse planner and the reference corpus. We also plan to correlate these metrics with human evaluation as proposed by Reiter and Sripada (2002).

## 5   Conclusions and future work

This paper has presented a machine learning approach to the fact-ordering subtask of discourse planning. We have decomposed the problem into a sequence of multi-class classification stages, where each stage selects the fact to be placed at the corresponding position. Experiments performed using the C4.5 and *k*-NN learning algorithms indicate that our method performs significantly better than both a sequence of simple majority classifiers and a set of manually constructed ordering rules.

Our method can be used with any content determination module that selects a fixed number of facts per document and user type, and gives rise to planners that can be easily retrained for other similar application domains, where sample manually ordered sequences of facts can be obtained. Compared to approaches that employ manually constructed rules, our method has the advantage that it does not require language technology expertise, and, hence, can be used to construct authoring tools that will allow domain experts to control the order of the facts in the generated documents. Furthermore, unlike previous machine learning approaches, our method does not interleave fact ordering with content determination.

As already mentioned, we plan to move towards learning richer discourse plans, which apart from ordering information will also include rhetorical relations, although our experience so far indicates that even just ordering the facts in a natural way can lead to quite acceptable texts. We are currently investigating a more flexible representation that will not be limited by a fixed number of facts per page and, apart from the absolute order of facts, will take into account the relative ordering between facts (e.g., by using n-grams). Further work is planned in order to devise better evaluation measures, and improve the performance of our planners by considering other learning algorithms.

## References

Aha D., and Kibler D. 1991. Instance-based learning algorithms. *Machine Learning*, 6:37–66.

Androutsopoulos I., Kokkinaki V., Dimitromanolaki A., Calder J., Oberlander J. and Not E. 2001. Generating multilingual personalized descriptions of museum exhibits – the M-PIRO project. In *Proceedings of the 29th Conference on Computer Applications and Quantitative Methods in Archaeology*, Gotland, Sweden.

Androutsopoulos I., Spiliotopoulos D., Stamatakis K., Dimitromanolaki A., Karkaletsis V. and Spyropoulos C.D. 2002. Symbolic authoring for multilingual natural language generation. In *Proceedings of the 2nd Hellenic Conference on Artificial Intelligence (SETN-02)*, Thessaloniki, Greece.

Bangalore S. and Rambow O. 2000. Exploiting a probabilistic hierarchical model for generation. In *Proceedings of the 18th International Conference on Computational Linguistics (COLING 2000)*, Saarbrucken, Germany.

Barzilay R., Elhadad N. and McKeown K. 2002. Inferring Strategies For Sentence Ordering In Multidocument News Summarization. *Journal of Artificial Intelligence Research*, 17: 35-55.

Duboue P and McKeown K. 2002. Content Planner Construction via Evolutionary Algorithms and a Corpus-based Fitness Function. In *Proceedings of the 2nd International Natural Language Generation Conference (INLG'02)*, New York, USA, pp. 89-96.

Duboue P. and McKeown K. 2001. Empirically estimating order constraints for content planning in generation. In *Proceedings of the 39th Annual Meeting of the Association for Computational Linguistics (ACL-2001)*, Toulouse, France, pp. 172-179.

Hovy E. 1993. Automated Discourse Generation Using Discourse Structure Relations. *Artificial Intelligence*, 63(1–2):341–386.

A. Isard, J. Oberlander, I. Androutsopoulos and C. Matheson. 2003. "Speaking the Users' Languages". *IEEE Intelligent Systems*, 18(1):40-45.

Kan M. and McKeown K. 2002. Corpus-trained text generation for summarization. In *Proceedings of the 2nd International Natural Language Generation Conference (INLG'02)*, New York, USA, pp. 1-8.

Karamanis N. and Manurung H. M. 2002. Stochastic Text Structuring using the Principle of Continuity. In *Proceedings of the 2nd International Natural Language Generation Conference (INLG'02)*, New York, USA, pp. 81-88.

Langkilde I and Knight K. 1998. Generation that Exploits Corpus-Based Statistical Knowledge. In

*Proceedings of the 36th Annual Meeting of the Association for Computational Linguistics and 17th International Conference on Computational Linguistics (COLING-ACL 1998)*, Montreal, Canada, pp. 704–710.

Malouf R. 2000. The order of prenominal adjectives in natural language generation. In *Proceedings of the 38th Annual Meeting of the Association for Computational Linguistics (ACL-00)*, Hong Kong, pp. 85-92.

Mann W. and Thompson S. 1988. Rhetorical structure theory: towards a functional theory of text organization. *Text*, 3:243–281.

McKeown K. 1985. Discourse strategies for generating natural language text. *Artificial Intelligence*, 27:1–42.

Marcu D. 1997. From local to global coherence: A bottom-up approach to text planning. In *Proceedings of the 14th National Conference on Artificial Intelligence*, Providence, Rhode Island, pp. 629-635.

Mellish C., Knott A., Oberlander J. and O' Donnell M. 1998. Experiments using stochastic search for text planning. In *Proceedings of the 9th International Workshop on Natural Language Generation*, Ontario, Canada, pp. 97-108.

O'Donnell M., Mellish C., Oberlander J. and Knott A. 2001. ILEX: An Architecture for a Dynamic Hypertext Generation System. *Natural Language Engineering*, 7(3):225–250.

Oh A. and Rudnicky A. 2000. Stochastic language generation for spoken dialogue systems. In *Proceedings of the ANLP/NAACL 2000 Workshop on Conversational Systems*, Seattle, USA, pp. 27–32.

Poesio M., Henschel R. and Kibble R. 2000. Statistical NP generation: a first report. In *Proceedings of the ESSLLI Workshop on NP Generation*, Utrecht, Netherlands.

Quinlan R. 1993. *C4.5: programs for machine learning*. Morgan Kaufmann, 302 p.

Rambow O. 1990. Domain Communication Knowledge. In *Proceedings of the 5th International Workshop on Natural Language Generation*, Dawson, PA.

Ratnaparkhi A. 2000. Trainable methods for surface natural language generation. In *Proceedings of the 6th Applied Natural Language Processing Conference and the 1st Meeting of the North American Chapter of ACL (ANLP-NAACL 2000)*, Seattle, USA, pp. 194–201.

Reiter E. and Dale R. 2000. *Building natural language generation systems*. Cambridge University Press, England, 248 p.

Reiter E. and Sripada S. 2002. Should Corpora Texts Be Gold Standards for NLG? In *Proceedings of the 2nd International Natural Language Generation Conference (INLG'02)*, New York, USA, pp. 97-104.

Shaw J. and Hatzivassiloglou V. 1999. Ordering among premodifiers. In *Proceedings of the 37th Annual Meeting of the Association for Computational Linguistics (ACL-99)*, College Park, Maryland, pp. 135–143.

Varges S. and Mellish C. 2001. Instance-based natural language generation. In *Proceedings of the 2nd Meeting of the North American Chapter of ACL (NAACL-2001)*, Carnegie Mellon University, Pittsburgh, PA.

Walker M. 2000. An application of reinforcement learning to dialogue strategy selection in a spoken dialogue system for email. *Journal of Artificial Intelligence Research*, 12:387–416.

Walker M., Rambow O. and Rogati M. 2001. SPoT: a trainable sentence planner. In *Proceedings of the 2nd Meeting of the North American Chapter of the ACL (NAACL-2001)*, Carnegie Mellon University, Pittsburgh, PA.

Witten I. and Frank E. 1999. *Data mining: practical machine learning tools and techniques with Java implementations*. Morgan Kaufmann, 416 p.